%% file: acl_latex.tex
\pdfoutput=1
\documentclass[11pt]{article}
\usepackage{acl}
\usepackage{times}
\usepackage{latexsym}
\usepackage[T1]{fontenc}
\usepackage[utf8]{inputenc}
\usepackage{microtype}
\usepackage{inconsolata}
\usepackage{graphicx}
\usepackage{amsmath}
\usepackage{amssymb}
\usepackage{booktabs}
\usepackage{multirow}
\usepackage{xcolor}
\usepackage{subcaption}
\usepackage{url}

\newcommand{\Likab}{\textsc{Likability}}
\newcommand{\Anime}{\textsc{AnimeScore}}
\newcommand{\Utmos}{\textsc{UTMOS}}
\newcommand{\Arousal}{\textsc{VAD-Arousal}}

\title{When Does Predictor-Based RL Align with Human Perception?\\
A Study of Subjective Rewards in Codec-Based Speech Language Models}

\author{Joonyong Park, Jerry Li \\
  Spellbrush \\
  \texttt{jyjoon97@gmail.com} \quad \texttt{jerry@sizigistudios.com}}

\begin{document}
\maketitle

\begin{abstract}
Codec-based text-to-speech (TTS) models make language-model post-training applicable to speech generation, yet it remains unclear when learned perceptual predictors can serve as reinforcement-learning rewards without losing human alignment.
We study this question with Group Relative Policy Optimization (GRPO) using learned rewards for anime-like speaking style, naturalness, likability, and arousal.
To prevent optimization through transcript drift, we introduce a character error rate (CER) zone constraint that applies the perceptual reward only when automatic speech recognition (ASR)-based CER stays below a threshold, and compare GRPO with Best-of-N reranking under the same gate.
Across single-reward runs, each reward primarily improves its target metric, showing that, in our settings, subjective predictors are not interchangeable quality surrogates.
Multi-rater A/B tests reveal uneven human transfer, while reward-gap analysis separates average transfer from within-axis calibration: signed reward gaps significantly predict listener choices both pooled and on each axis, whereas residual CER gaps do not.
In human A/B tests, listeners do not clearly prefer GRPO over Best-of-8, suggesting that GRPO amortizes reward-selected behavior into the policy rather than uniformly outperforming reranking.
These results support analyzing subjective speech rewards as predictor–axis–base tuples (predictor, target attribute, base model) and provide practical diagnostics before multi-reward post-training.
\end{abstract}

\vspace{-1mm}\section{Introduction}
\label{sec:intro}

Modern text-to-speech (TTS) systems increasingly generate speech through \emph{discrete acoustic tokens}.
A neural audio codec first converts a waveform into a sequence of discrete codes, and a speech language model then predicts these codes autoregressively from text and, in zero-shot settings, a short reference audio prompt.
The generated token sequence is finally decoded back into a waveform.
This codec-based formulation underlies recent systems~\citep{wang2023valle, borsos2023audiolm, du2024cosyvoice, ye2025llasa}, and makes speech generation resemble conditional language modeling over acoustic tokens.
As a result, post-training methods developed for large language models (LLMs), including reinforcement learning (RL) from automatic or learned rewards, can now be applied to TTS policies.
Recent RL work for TTS has shown that such methods can improve automatically measurable properties of generated speech, including transcription fidelity, target-text likelihood, speaker similarity, duration control, and prosodic stability~\citep{liu2025grpotts, li2026dmospeech2, zhong2025multireward, gao2025diffro}.

However, many useful speech-generation objectives are not verifiable in the same way as transcription accuracy.
Naturalness, expressiveness, likability, affective intensity, and domain-specific speaking style are perceptual attributes: they are ultimately defined by listener judgments rather than by a symbolic target string.
A natural approach is to approximate such listener-defined attributes with learned perceptual predictors, such as mean opinion score (MOS) or speech-quality predictors, and use their scalar outputs as training rewards~\citep{saeki2022utmos, mittag2021nisqa, chen2024dlpo}.
The difficulty is that a learned perceptual predictor is only a proxy~\citep{ziegler2019finetuning, gao2023scaling}.
A speech policy can increase the predictor score by exploiting predictor blind spots, producing outputs that are high-scoring but unintelligible, unstable, artifact-laden, or not actually preferred by listeners.
This risk is especially acute in speech because generated waveforms must jointly satisfy transcription fidelity, acoustic quality, speaker similarity, naturalness, and prosodic or stylistic appropriateness~\citep{chen2024valle2, mittag2021nisqa}.
Existing TTS-RL systems therefore often rely on verifiable or automatically computed reward components, such as character error rate (CER), word error rate (WER), negative log-likelihood (NLL), speaker similarity, duration, entropy, and rule-based prosody rewards~\citep{liu2025grpotts, li2026dmospeech2, zhong2025multireward, gao2025diffro}.
These rewards are useful for stabilizing generation, but they do not answer when a learned subjective predictor can be optimized while remaining aligned with listener judgments.

\begin{figure}[t]
  \centering
\includegraphics[width=\columnwidth]{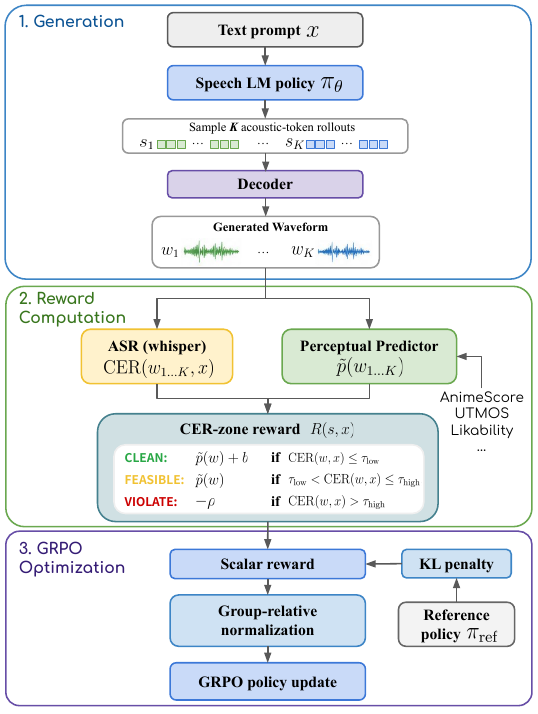}
\caption{Overview of constrained perceptual GRPO.}
\label{fig:method_overview}
  \vspace{-1.5mm}
\end{figure}

This paper asks when predictor-based RL can move a codec speech language model along a subjective reward axis while preserving human-perceptual alignment.
We study this question with group relative policy optimization (GRPO) using learned perceptual predictors as rewards.
A perceptual predictor evaluates acoustic or stylistic properties rather than lexical content, so an unconstrained policy can trade transcription fidelity for style.
We therefore gate each reward by the CER of an automatic transcript of the generated speech (Figure~\ref{fig:method_overview}).
A clip with CER at most 0.10 receives the predictor score plus a small bonus (CLEAN), a clip with CER between 0.10 and 0.30 receives the predictor score alone (FEASIBLE), and a clip with CER above 0.30 receives a fixed penalty regardless of its perceptual score (VIOLATE).
We call these three cases \emph{CER zones}.
Without this gate, the policy increases its predictor score partly by drifting from the target transcript. Median CER rises on the reward axes, whereas the gated policy achieves an equal or larger target gain with lower CER (\S\ref{sec:best_of_n}).
We use a hard gate rather than a weighted CER penalty because no perceptual advantage can compensate for unintelligible speech.
We also compare GRPO with Best-of-$N$ reranking under the same reward gate, separating inference-time reward selection from policy-level movement.
Because different reward settings vary not only in perceptual target but also in predictor architecture, training data, score scale, and base-model distribution, we analyze each setting as a \emph{predictor--axis--base tuple}, for example the UTMOS22 predictor applied to the naturalness axis on the Llasa-1B-Multilingual base, rather than attributing success or failure to the subjective axis alone.
We use anime-like speaking style as our primary case study because it is readily recognizable to listeners, making preference modeling and pairwise A/B evaluation feasible, yet difficult to specify with symbolic rules.
The other two main axes, naturalness and likability, are domain-general.

Our study combines three levels of evidence.
First, we compare single-reward GRPO runs under the same CER-zone scaffold to determine whether each reward produces a targeted machine-level shift or merely a generic quality change.
Second, we use multi-rater A/B tests to evaluate whether those machine-level shifts are perceived by listeners.
Third, we analyze reward signal quality using diagnostics such as reward-gap calibration, base-output score spread, domain match, and within-zone signal strength.
This design allows us to distinguish three phenomena that are often conflated: whether a subjective predictor can be optimized, whether the resulting policy shift remains human-aligned, and whether the reward is suitable for inclusion in future multi-reward post-training.

Our main contributions are:
\begin{itemize}
    \item \textbf{A controlled training-and-evaluation scaffold for subjective speech rewards.}
    We instantiate GRPO for learned perceptual predictors under a CER-zone hard constraint, compare it with Best-of-$N$ reranking to separate reward-based sample selection from policy-level movement, test whether listeners prefer one over the other, and evaluate both first-shot behavior and CER-retry behavior used for human evaluation.
    \item \textbf{A multi-rater human study of predictor--axis--base tuples.}
    We show that machine-level reward gains transfer unevenly to listeners: some tuples yield strong human-aligned shifts, some yield only modest transfer, and some fail on average despite high-confidence successes.
    \item \textbf{Diagnostics for reward signal quality.}
    We evaluate reward-gap calibration, base-output spread, domain match, and within-zone signal strength as diagnostics for deciding which subjective rewards are suitable for RL or future multi-reward post-training.
\end{itemize}

Together, these results characterize not merely whether subjective predictors can be optimized, but when, for the predictor--axis--base tuples we study, predictor-based RL remains aligned with human perception in codec-based speech language models.
We release code, prompts, generated audio samples, reward scores, and anonymized human-evaluation votes at https://github.com/sizigi/animeGRPO.\footnote{Audio demo: https://sizigi.github.io/animeGRPO/}

\vspace{-0mm}\section{Related Work}
\label{sec:related}

\vspace{-1mm}\paragraph{Codec-based speech language models.}
Codec-based speech language models formulate TTS as autoregressive generation over discrete acoustic tokens produced by neural audio codecs such as SoundStream and EnCodec~\citep{zeghidour2021soundstream, defossez2023encodec}.
Systems such as VALL-E, AudioLM, CosyVoice, and Llasa then generate these tokens from text and optional acoustic context~\citep{wang2023valle, borsos2023audiolm, du2024cosyvoice, ye2025xcodec, ye2025llasa}.
Such codecs are themselves trained with multiple differentiable objectives, including reconstruction, adversarial, and quantization losses; we instead keep the codec fixed and post-train the token-generating language model with learned, non-differentiable perceptual rewards.

\vspace{-1mm}\paragraph{RL and preference optimization for TTS.}
Group Relative Policy Optimization (GRPO) was introduced as a critic-free variant of Proximal Policy Optimization that estimates advantages from group scores~\citep{shao2024deepseekmath}.
Recent work has begun to apply GRPO, preference optimization, or reward-based fine-tuning to TTS.
GRPO for TTS has optimized automatic speech recognition (ASR)-derived rewards such as CER and NLL~\citep{liu2025grpotts}.
DMOSpeech 2 applies GRPO to duration prediction with speaker similarity and WER-based rewards~\citep{li2026dmospeech2}.
Multi-Reward GRPO combines intelligibility and speaker-similarity objectives with rule-based rewards for length, decoding stability, and prosody alignment~\citep{zhong2025multireward}.
DiffRO optimizes neural codec language models with differentiable reward prediction from speech tokens~\citep{gao2025diffro}.
SpeechAlign studies preference-based optimization for speech generation using direct preference optimization (DPO), proximal policy optimization (PPO), and Best-of-$N$ selection~\citep{zhang2024speechalign}.
These studies establish that reward-based post-training can improve TTS, but they primarily focus on verifiable, automatically computed, or task-specific reward components.
Those that report human evaluations use aggregate measures such as MOS or A/B win rates; to our knowledge, none examines at the vote level whether the size of a subjective reward gap predicts individual listener choices.

\vspace{-1mm}\paragraph{Learned perceptual predictors.}
A long line of speech evaluation work aims to predict subjective listener judgments automatically.
Early neural MOS predictors such as MOSNet model human naturalness ratings for converted or synthesized speech~\citep{lo2019mosnet}.
More recent non-intrusive quality predictors, including UTMOS and NISQA, estimate naturalness or multidimensional speech quality without reference audio~\citep{saeki2022utmos, mittag2021nisqa}.
Such predictors turn perceptual judgments into scalar model outputs and can therefore serve as proxy objectives for generation or post-training~\citep{chen2024dlpo}.
However, a predictor score is not equivalent to a human judgment: learned predictors can be miscalibrated, out-of-domain, insensitive to relevant perceptual differences, or vulnerable to overoptimization.
This motivates evaluating not only whether a perceptual score increases, but whether the increase corresponds to listener preference.

\vspace{-1mm}\paragraph{Reward overoptimization and calibration.}
Learned rewards are known to suffer from overoptimization: a policy can obtain high proxy reward while degrading the true human objective~\citep{gao2023scaling}.
Common mitigations include Kullback--Leibler (KL) regularization, constrained optimization, improved reward modeling, and inference-time reranking~\citep{ziegler2019finetuning, achiam2017cpo}.
In speech, this problem is compounded by the need to satisfy multiple coupled constraints, including intelligibility, speaker consistency, acoustic quality, and style.
This makes calibration between machine reward differences and human-perceived differences particularly important.
This concern applies both to policy optimization and to inference-time Best-of-$N$ selection, since both can overoptimize a proxy reward~\citep{gao2023scaling}.

\vspace{-1mm}\section{Constrained Perceptual GRPO}
\label{sec:method}

\vspace{-1mm}\subsection{Problem setup}

Given a text prompt $x$, the codec speech language model samples a discrete acoustic-token sequence
$s=(s_1,\ldots,s_T)\sim\pi_\theta(\cdot\mid x)$.
A codec decoder maps $s$ to a waveform $w=\mathrm{Dec}(s)$.
Perceptual predictors are applied to the decoded waveform, but for brevity we write
$p(s) := p(\mathrm{Dec}(s))$.
The goal is to improve a target perceptual attribute while keeping the generated speech intelligible and close to a frozen reference policy $\pi_{\mathrm{ref}}$.

\vspace{-1mm}\subsection{CER-zone reward template}

Let $c(s,x)$ be the CER between an ASR transcript of $w=\mathrm{Dec}(s)$ and the input text $x$.
For a learned perceptual predictor score $p(s)$, we define
\begin{equation}
\label{eq:Rgeneric}
R(s,x)=
\begin{cases}
\tilde{p}(s)+b, & c(s,x) \le \tau_l,\\
\tilde{p}(s), & \tau_l < c(s,x) \le \tau_h,\\
-\rho, & c(s,x) > \tau_h.
\end{cases}
\end{equation}
The three cases correspond to CLEAN, FEASIBLE, and VIOLATE zones.
Here $\tilde{p}(s)$ is the normalized non-negative predictor score, $b$ is the CLEAN-zone bonus, and $\rho$ is the fixed VIOLATE penalty.
In all experiments, we set $\tau_l=0.10$, $\tau_h=0.30$, $b=0.5$, and $\rho=1.0$.
This hard-zone design prevents a high perceptual score from numerically compensating for a transcription failure.
The predictor-specific definitions of $\tilde{p}(s)$ are given in \S\ref{sec:reward_predictors}.

\vspace{-1mm}\subsection{GRPO optimization and stopping}
\label{sec:stopping}

We use GRPO that estimates advantages by comparing multiple rollouts for the same prompt.
For each prompt, we sample $K=4$ rollouts and compute a scalar reward $r_i$ by subtracting an adaptive in-reward KL penalty from $R(s_i,x)$.
We then normalize rewards within the rollout group as
\[
\hat A_i=(r_i-\mu_r)/(\sigma_r+\epsilon).
\]
Implementation details, including the \texttt{verl} and vLLM rollout setup, are reported in Appendix~\ref{app:hparams}.

We do not select checkpoints by the raw perceptual predictor score alone.
Instead, we select the checkpoint that maximizes constraint-aware validation reward while monitoring CER violations and KL drift.
This is important because larger predictor scores or reward gaps can arise from proxy overoptimization, transcript drift, or predictor blind spots.
Selected checkpoints and full hyperparameters are reported in Appendix~\ref{app:hparams}, and KL trajectories are reported in Appendix~\ref{app:kl_csv}.

\vspace{-1mm}\section{Experimental Setup}
\label{sec:exp}

\vspace{-1mm}\subsection{Base speech model}
\label{sec:base_models}

We use Llasa as the codec-based TTS backbone and XCodec2 as the acoustic tokenizer and waveform decoder~\citep{ye2025xcodec,ye2025llasa}.
We used a public checkpoint\footnote{https://huggingface.co/HKUSTAudio/Llasa-1B-Multilingual}, a multilingual variant trained with Emilia and Multilingual LibriSpeech (MLS), which provide Japanese-containing in-the-wild multilingual speech and multilingual read-speech data, respectively~\citep{he2024emilia,pratap2020mls}.
Our GRPO implementation builds on \texttt{verl}~\citep{sheng2024verl}.
All main runs update the full actor parameters, and the KL reference policy is a frozen copy of the same base checkpoint.
The auxiliary English \Anime{} experiment is reported in Appendix~\ref{app:anime_en} and is used only as supporting evidence for cross-language behavior.
\vspace{-1mm}\subsection{Reward predictors}
\label{sec:reward_predictors}

We use three learned perceptual predictors in the main experiments and one additional arousal predictor for a training-only diagnostic in Appendix~\ref{app:arousal_failure}.
\Anime{}\footnote{https://github.com/sizigi/animescore} is a pairwise-preference-trained anime-likeness predictor~\citep{park2026animescore}, using a WavLM-base encoder, temporal modeling, and a RankNet-style ranking head.
Its raw output is a signed score, which we normalize as
\[
\tilde{p}(s)=\max(0,p(s)+3.0).
\]

\Utmos{} is used off-the-shelf from UTMOS22-strong\footnote{https://github.com/sarulab-speech/UTMOS22}~\citep{saeki2022utmos}.
It predicts a MOS-like naturalness score on $[1,5]$, and we set $\tilde{p}(s)=p(s)/5$.

\Likab{} is a likability predictor trained for this study on CocoNut-Humoresque~\citep{suda2025coconut_humoresque}.
It uses a pretrained WavLM encoder\footnote{https://huggingface.co/microsoft/wavlm-base-plus}, and its raw score is the expected class value on $[1,6]$.
We use the original CocoNut-Humoresque split and train with class-weighted cross-entropy rather than regression to preserve score spread for GRPO.
For reward computation, we set
\[
\tilde{p}(s)=\frac{p(s)-1}{5}.
\]

We additionally use an off-the-shelf MSP-Dim arousal predictor only for the training-dynamics negative case discussed in \S\ref{sec:diagnostic_within_zone}; details are in Appendix~\ref{app:arousal_failure}.

For the CER gate, we transcribe generated waveforms with Whisper large-v3~\citep{radford2023whisper} and compute CER against the canonical written prompt.
No additional language-specific text normalization is applied.
Additional reward-predictor training and implementation details are provided in Appendix~\ref{app:reward_servers}.

\vspace{-1mm}\subsection{Training and evaluation prompts}
\label{sec:prompts}

GRPO training uses 900 Japanese Wikipedia-derived prompts.
All main reward-axis runs use the same prompt training set to isolate the reward predictor as the intended varying factor.
Checkpoint selection uses a disjoint 100-prompt validation set; the auxiliary English run trains on 900 English prompts (Appendix~\ref{app:anime_en}).

The main evaluation set contains 50 held-out Japanese prompts and is disjoint from GRPO training, validation, and reward-model training data.
It is partitioned into five groups: emotional, anime-stylized, neutral, long-form narrative, and linguistically challenging prompts.
Auxiliary English prompts are idiomatic English counterparts of the Japanese test items with the same group structure, written for English listeners rather than translated literally; they are described in Appendix~\ref{app:anime_en}.

\vspace{-1mm}\subsection{Decoding and evaluation protocol}
\label{sec:clean_protocol}

We distinguish two evaluation modes.
\textbf{First-shot evaluation} uses a single stochastic generation with fixed seed and no filtering or regeneration.
We use this setting to measure the model's unfiltered behavior, including the raw violation rate, defined as the fraction of outputs with CER $>0.30$.

\textbf{CER-retry evaluation} is used to prepare audio for human-level evaluation.
Its purpose is to reduce obvious transcript-failure confounds while avoiding asymmetric post-hoc filtering.
For both base and GRPO systems, we first generate with fixed seed.
If the output has CER $>0.30$, we re-generate with up to 12 retry seeds and select the first candidate with CER $\le 0.30$.
If no candidate satisfies the threshold, we keep the lowest-CER candidate.
This rule is applied symmetrically to both sides of each pair.
All 50 prompts are retained, including residual CER violators after all retries, to avoid post-hoc filtering by a metric tied to the RL scaffold.

\vspace{-1mm}\subsection{Evaluation metrics}
\label{sec:metrics}

We evaluate each reward-axis run with both machine-level and human-level evaluations.
\vspace{-1mm}\paragraph{Machine-level evaluation.}
For each system, we generate speech on the 50-prompt held-out test set and report changes from the base model.
For the target perceptual axis, we report the corresponding predictor delta; we also score each generated sample with the other predictors to measure cross-axis side effects.
To measure transcription fidelity, we compute CER between the input text and a Whisper transcript of the generated waveform, and report both mean and median CER.
We also report the violation rate, defined as the fraction of generated samples with CER $>0.30$.
For first-shot evaluation, this violation rate is measured from a single seed generation.
For CER-retry evaluation and human-evaluation audio, it is measured after applying the symmetric retry protocol described in \S\ref{sec:clean_protocol}.

We also compare GRPO with Best-of-$N$ reranking.
For Best-of-$N$, we sample $N$ candidates from the base model, score each candidate with the same CER-gated reward used for GRPO, and select the highest-scoring candidate.
In the human study, the Best-of-8 side is this candidate without CER retry.
This comparison tests whether reward-selected samples already exist in the base model's support, while GRPO tests whether such selection behavior can be amortized into the policy.

\vspace{-1mm}\paragraph{Human-level evaluation.}
We conduct pairwise listening tests through Lancers\footnote{https://www.lancers.jp}, a Japanese crowdsourcing platform.
The main human study covers three reward axes: \Anime{}, \Utmos{}, and \Likab{}.
Each axis contains 50 paired items, each item receives 5 independent ratings, and each axis is rated by 10 distinct Japanese listeners with no listener overlap between axes.
The auxiliary English \Anime{} study is reported in Appendix~\ref{app:anime_en}.

Each pair is presented as A/B audio with randomized side assignment, and listeners are not told which side corresponds to base, GRPO, or Best-of-$N$.
For \Anime{}, listeners choose the clip that sounds more anime-like, defined as voice-actor-like speaking style and performance.
For \Utmos{}, listeners choose the clip with better naturalness and audio quality, defined as fewer artifacts, noise, and distortions.
For \Likab{}, listeners choose the clip that sounds more pleasing or comfortable in voice and speaking manner.
A separate skip path exists for problematic clips, but those are excluded from win-rate calculations.

At the item level, a system wins an item if more than half of raters prefer that system.
Human win rate (HWR) is the fraction of items for which the listener majority prefers the target system.
Machine win rate (MWR) is the fraction of items for which the reward predictor prefers the target system.
Agreement is the fraction of items for which the machine winner and the human-majority winner are identical.
Unless otherwise stated, HWR, MWR, and agreement are item-level metrics: when individual listener votes are analyzed, we explicitly label the metric as Vote HWR.
We report item-level majority results as the primary human-evaluation metric, with Wilson 95\% confidence intervals in Appendix~\ref{app:human_stats}.

\begin{table}[t]
\centering
\small
\setlength{\tabcolsep}{3.0pt}
\begin{tabular}{@{}lrrrr@{}}
\toprule
Reward & $\Delta$CER & $\Delta$AS & $\Delta$Likab. & $\Delta$UTMOS \\
\midrule
\Anime{} & $-0.030$ & {\boldmath$+1.353$} & $+0.075$ & $-0.037$ \\
\Likab{} & $-0.018$ & $+0.029$ & {\boldmath$+0.167$} & $+0.090$ \\
\Utmos{} & $-0.030$ & $-0.072$ & $+0.140$ & {\boldmath$+0.485$} \\
\bottomrule
\end{tabular}
\caption{Cross-axis mean objective shifts between base and Zone-CER GRPO outputs.
Values are changes from the corresponding base outputs.}
\label{tab:heatmap4}
\end{table}

\vspace{-1mm}\section{Machine-Level Behavior and Baselines}
\label{sec:single_reward}

\vspace{-1mm}\subsection{Cross-axis specificity and machine-level baselines}
\label{sec:cross_axis}

We first examine whether subjective rewards induce targeted machine-level shifts, whether the CER-zone scaffold improves the reward--intelligibility trade-off, and whether policy optimization provides benefits beyond inference-time reranking.

Table~\ref{tab:heatmap4} evaluates each single-reward GRPO run on all reward axes.
For each evaluation metric, the largest mean gain is obtained by optimizing its corresponding predictor, indicating that subjective rewards mainly induce axis-specific movement rather than a generic quality improvement.
This diagonal pattern shows that, among the predictors we study, no single subjective reward is a universal surrogate for all desired speech properties.

\vspace{-1mm}\subsection{Policy optimization versus Best-of-\texorpdfstring{$N$}{N} reranking}
\label{sec:best_of_n}

We then compare three ways of using the same reward signal:
Best-of-$N$ does not update the policy: it samples $N$ candidates from the base model, scores them with the CER-gated reward, and serves the highest-scoring candidate.
Target-only and Zone-CER are policy-trained systems.
Target-only optimizes the perceptual predictor without the CER-zone constraint, while Zone-CER uses the full reward in Eq.~\eqref{eq:Rgeneric}.

Table~\ref{tab:machine_baselines} reports method-specific machine behavior: Base, Target-only, and Zone-CER are evaluated from one policy sample, while Best-of-$N$ selects the highest-scoring candidate among $N$ base samples.
The Best-of-$N$ rows show that reward-selected samples already exist in the base model's support.
This machine-level pattern is consistent with prior work on reranking and reward overoptimization~\citep{gao2023scaling, zhang2024speechalign}. However, proxy reward scores alone cannot determine whether listeners prefer one method over the other, which we test directly in \S\ref{sec:human_bon}.
On \Anime{}, Zone-CER produces a larger target shift than Best-of-8, but Best-of-8 has slightly lower median CER and fewer violations.
On \Utmos{}, Best-of-8 and Zone-CER achieve almost identical target gains, and Zone-CER has slightly lower median CER and fewer violations.
On \Likab{}, Zone-CER gives the largest target gain, but the gain is small and comes with higher median CER than Best-of-8, although with fewer violations.
Thus, GRPO is not uniformly better than reranking; the value of policy optimization depends on the reward axis and deployment trade-off.
The Target-only rows additionally isolate the role of the CER-zone constraint: removing the CER zone worsens median CER and violation rate, while Zone-CER preserves or improves the target gain relative to Target-only with the largest differences on \Anime{} and UTMOS.

\label{sec:ablation}

\begin{table}[t]
\centering
\small
\setlength{\tabcolsep}{3.0pt}
\begin{tabular}{@{}llrrr@{}}
\toprule
Axis & Method & $\Delta$Target & CER med. & Viol.(\%) \\
\midrule
\Anime{} & Base      & $0.00$  & $0.058$ & {$12.0$} \\
         & Best-of-4 & $+0.83$ & $0.064$ & $10.0$ \\
         & Best-of-8 & $+1.21$ & $0.070$ & $10.0$ \\
\cmidrule(lr){2-5}
         & Target-only & $+1.08$ & $0.087$ & $24.0$ \\
         & Zone-CER    & $+1.35$ & {$0.073$} & {$12.0$} \\
\midrule
\Utmos{} & Base      & $0.00$  & $0.058$ & {$12.0$} \\
         & Best-of-4 & $+0.32$ & $0.052$ & $10.0$ \\
         & Best-of-8 & $+0.47$ & $0.055$ & $10.0$ \\
\cmidrule(lr){2-5}
         & Target-only & $+0.25$ & {$0.082$} & {$16.0$} \\
         & Zone-CER    & $+0.49$ & {$0.053$} & $\phantom{0}6.0$ \\
\midrule
\Likab{} & Base      & $0.00$  & $0.058$ & {$12.0$} \\
         & Best-of-4 & $+0.12$ & $0.046$ & $10.0$ \\
         & Best-of-8 & $+0.14$ & $0.043$ & $10.0$ \\
\cmidrule(lr){2-5}
         & Target-only & $+0.15$ & $0.119$ & $16.0$ \\
         & Zone-CER    & $+0.17$ & $0.098$ & $\phantom{0}6.0$ \\
\bottomrule
\end{tabular}
\caption{Machine-level comparison across reward axes.
$\Delta$Target denotes mean objective shifts depending on the axis ($\Delta$AS, $\Delta$UTMOS, or $\Delta$Likab.).
Base, Target-only, and Zone-CER use a single first-shot sample; Best-of-$N$ selects the highest-scoring candidate among $N$ base samples.}
\label{tab:machine_baselines}
\end{table}

\begin{table}[t]
\centering
\small
\setlength{\tabcolsep}{3.0pt}
\begin{tabular}{@{}lrrrr@{}}
\toprule
Axis & HWR & MWR & Agree & $\Delta$Target \\
\midrule
\multicolumn{5}{@{}l}{\textbf{GRPO vs Base}} \\
\midrule
\Anime{} & $80.0$ & $88.0$ & $88.0$ & $+1.24$ \\
\Utmos{} & $62.0$ & $74.0$ & $80.0$ & $+0.46$ \\
\Likab{}& $36.0$ & $56.0$ & $76.0$ & $+0.17$ \\
\Anime{} EN$^\dagger$ & $70.0$ & $66.0$ & $72.0$ & $+0.50$ \\
\midrule
\multicolumn{5}{@{}l}{\textbf{GRPO vs Best-of-8}} \\
\midrule
\Anime{} & $52.0$ & $50.0$ & $74.0$ & $+0.07$ \\
\Utmos{} & $46.0$ & $36.0$ & $68.0$ & $-0.01$ \\
\Likab{} & $48.0$ & $32.0$ & $62.0$ & $+0.03$ \\
\bottomrule
\end{tabular}
\caption{Human alignment results for GRPO against Base and Best-of-8. All HWR, MWR, and Agree values are item-level percentages over 50 paired items using majority vote from 5 raters per item. $\Delta$Target is computed as the mean GRPO target score minus the comparison system's target score.
$^\dagger$Auxiliary English \Anime{} study on 50 English items, rated by the same Japanese-native listener pool (Appendix~\ref{app:anime_en}).}
\label{tab:human_combined}
\end{table}

\vspace{-1mm}\section{Human Evaluation}
\label{sec:human}

\vspace{-1mm}\subsection{Axis-level transfer}
\label{sec:axis_transfer}

Table~\ref{tab:human_combined} shows that predictor-level gains transfer unevenly to listeners.
\Anime{} is the strongest positive case: the listener majority prefers the GRPO output on most items, and item-level machine--human agreement reaches 88\%.
\Utmos{} shows high machine--human agreement, but only a modest aggregate human preference shift.
\Likab{} is the main negative average-transfer case under this predictor--axis--base tuple: humans prefer the base overall. However, the reward-gap analysis below shows that this average failure masks a calibrated high-confidence region.
The auxiliary English \Anime{} study shows a similar trend (HWR 70.0, agreement 72.0), but because it uses the same Japanese-native listener pool, we treat it as supporting rather than primary evidence.
These results separate two notions of transfer.
Axis-level transfer asks whether the optimized system is preferred on average, while reward-gap calibration asks whether score differences explain item-level listener choices.
Under this distinction, \Anime{} is the clearest average-transfer success, whereas \Likab{} and \Utmos{} provide the clearest evidence of within-axis reward-gap calibration.

\vspace{-1mm}\subsection{Comparison between GRPO and Best-of-\texorpdfstring{$N$}{N} in human preference}
\label{sec:human_bon}

Table~\ref{tab:human_combined} also compares GRPO with Best-of-8 on each reward axis.
Across all three axes, human preference is near chance: HWR is 52.0 for \Anime{}, 46.0 for \Utmos{}, and 48.0 for \Likab{}.
Thus, listeners do not clearly prefer GRPO over Best-of-8.
Proxy reward scores cannot resolve this comparison, since both methods are evaluated under the same imperfect objective they optimize.
Under our evaluation protocol, the listening test establishes neither a clear preference for GRPO nor perceptual equivalence; the two methods often reach similar scores while differing in voice characteristics and speaking rate (Appendix~\ref{app:qual_bon}).
GRPO can reduce inference-time sampling and scoring by amortizing selection into the policy; training to the selected \Anime{} checkpoint took under two days on two H100 GPUs.
GRPO should therefore be interpreted as policy-level movement that amortizes reward-selected behavior rather than as a perceptual improvement over a strong Best-of-8 baseline.

\vspace{-1mm}\subsection{Reward-gap calibration}
\label{sec:gap_calibration}

Average win rates can hide whether a reward difference is perceptually meaningful.
We therefore test whether signed pairwise reward gaps predict listener choices beyond residual intelligibility differences.
Specifically, we pool the three main Japanese axes and the auxiliary English \Anime{} study, and fit a vote-level logistic regression predicting whether an individual listener chooses the GRPO side from the within-axis standardized signed reward gap $z_R$ and the within-axis standardized CER gap $z_C$, with axis fixed effects and item-clustered robust standard errors.
$z_R$ is positive when the target reward favors GRPO, and $z_C$ is positive when the GRPO side has lower CER.

The signed reward gap is a strong predictor of human choice (Table~\ref{tab:gap_logit}): a one-standard-deviation increase in reward gap in favor of GRPO increases the odds of choosing GRPO by $5.26\times$.
By contrast, the CER gap is not predictive in the pooled fit, and a Wald test rejects equality of the two slopes.
Thus, within the CER-retry evaluation regime, GRPO-side preference is better explained by signed reward advantage than by residual CER advantage.

Per-axis checks in Appendix~\ref{app:gap_regression}
show the same pattern on every axis, with smaller reward-gap slopes for the two \Anime{} studies (OR 4.05 and 4.17) and the largest slope for \Likab{} (OR 8.27).
Reward-gap calibration should therefore be read as a local confidence diagnostic within a predictor--axis--base tuple, not as the sole explanation of average transfer.
Full bin statistics are reported in Appendix~\ref{app:gap_bins}.

\vspace{-1mm}\subsection{Robustness to residual CER violations}
\label{sec:human_cer_robustness}

Although the CER-retry protocol reduces transcript failures, a small number of item pairs remain above the CER threshold after all retries.
We retain all 50 items to avoid post-hoc filtering by a metric tied to the RL objective.
Excluding residual violator pairs changes vote-level HWR by $+1.9$ pp for \Anime{}, $-2.1$ pp for \Utmos{}, and $-4.5$ pp for \Likab{}; the qualitative conclusions remain unchanged.

\begin{table}[t]
\centering
\small
\setlength{\tabcolsep}{2pt}
\begin{tabular}{@{}lrrrr@{}}
\toprule
Variable & $\beta$ (SE) & OR & 95\% CI & $p$ \\
\midrule
Reward gap $z_R$ & {$+1.660$ ($0.132$)} & {$5.26$} & {$[4.06,6.81]$} & $<0.001$ \\
CER gap $z_C$ & {$+0.142$ ($0.100$)} & {$1.15$} & {$[0.95,1.40]$} & {$0.16$} \\
\bottomrule
\end{tabular}
\caption{Vote-level logistic regression predicting whether a listener chooses the GRPO side, using axis fixed effects and item-clustered robust standard errors.
$z_R$ is the within-axis standardized signed reward gap in favor of GRPO, and $z_C$ is the within-axis standardized CER advantage of GRPO.}
\label{tab:gap_logit}
\end{table}

\begin{table}[t]
\centering
\small
\setlength{\tabcolsep}{3.0pt}
\begin{tabular}{@{}lrrr@{}}
\toprule
Axis & Mean & Std & Range \\
\midrule
\Anime{} & $-0.39$ & $1.53$ & $[-2.52,4.20]$ \\
\Utmos{} & $+3.08$ & $1.01$ & $[1.30,4.23]$ \\
\Likab{} & $+4.20$ & $0.44$ & $[2.78,4.58]$ \\
\bottomrule
\end{tabular}
\caption{Base predictor distributions on base model with test set.}
\label{tab:rewardscale}
\end{table}

\section{Discussion: Diagnostics for Predictor--Axis--Base Tuples}
\label{sec:diagnostics}

The human study shows that increasing a subjective predictor score is not sufficient for human-aligned transfer.
This is consistent with the broader observation in RL from human feedback that learned rewards are useful but imperfect proxies, and that optimizing them can diverge from the intended human objective~\citep{ziegler2019finetuning,gao2023scaling}.
In speech, this proxy gap is shaped not only by the reward model, but also by the perceptual axis and the base model distribution.
We therefore analyze each setting as a predictor--axis--base tuple. The diagnostics below are evidence-supported heuristics, not causal explanations or universal rules for reward success.
Because the predictors differ in architecture, training data, score scale, and domain match, comparisons across axes are confounded; we make no claim that one perceptual axis is intrinsically easier to optimize than another, and each diagnostic is checked within its own tuple.

\vspace{-1mm}\subsection{Reward-gap calibration}
\label{sec:diagnostic_gap}

The strongest diagnostic observed in our study is reward-gap calibration, but it should be distinguished from average transfer.
Instead of asking only whether the optimized system is preferred on average, reward-gap calibration asks whether larger signed reward advantages correspond to listener choices within a given predictor--axis--base tuple.
The regression analysis in \S\ref{sec:gap_calibration} supports this view at the pooled level: standardized signed reward gaps predict GRPO-side listener preference, whereas residual CER gaps do not.
The per-axis fits (Appendix~\ref{app:gap_regression}) show that this holds on every axis, yet \Anime{}, the strongest average-transfer case, has the weakest within-axis slope.
The optimization appears to move the distribution in a perceptually salient direction, while the magnitude of item-level reward gaps is less informative once many items already favor GRPO.

This diagnostic is especially useful for \Likab{}.
Although \Likab{} fails on average, its per-axis reward-gap slope is significant and the steepest, indicating that the predictor is informative in high-confidence regions.
The negative average result therefore should not be read as evidence that likability is intrinsically difficult to improve.
Rather, under this predictor--axis--base tuple, few items reach a reward-gap regime that is perceptually reliable.
In the binned analysis (Appendix~\ref{app:gap_bins}), 40 of the 50 \Likab{} items fall in the low-gap bin ($|\Delta| < 0.3$ on its raw scale), with 61.5\% machine--human agreement and 24.5\% GRPO preference, compared with 90--100\% agreement in the two larger-gap bins (8 and 2 items). Thus, the average-transfer failure is concentrated in this low-gap regime.

\vspace{-1mm}\subsection{Base-output spread}
\label{sec:diagnostic_spread}

A second screening signal is the predictor's score spread on base-model outputs.
If a predictor assigns nearly identical scores to naturally occurring base outputs, this may indicate limited resolution on the target base distribution.
Table~\ref{tab:rewardscale} is consistent with part of this pattern: \Anime{} has the widest spread and gives the strongest positive transfer, while \Likab{} has narrower spread and fails on average.
This across-prompt spread on raw scales is only a heuristic for the within-prompt signal that GRPO uses.
The \Arousal{} failure is discussed separately in Appendix~\ref{app:arousal_failure}.

\vspace{-1mm}\subsection{Within-zone signal under constraints}
\label{sec:diagnostic_within_zone}

The \Arousal{} run illustrates a constraint-specific failure mode.
A predictor can be meaningful as a standalone evaluator but still fail as an RL reward if its variation inside the feasible CER zone is too small relative to the constraint penalty.
In our \Arousal{} run, validation reward rose from $-0.34$ to $-0.05$ as the violation rate fell from 0.62 to 0.45, while validation arousal stayed within 0.61--0.64, a net change of $+0.014$ (Appendix~\ref{app:arousal_failure}).
Because no human A/B study was conducted, we treat this as a training-dynamics negative case rather than evidence about perceptual transfer.
These dynamics are consistent with weak within-zone arousal variation relative to the constraint penalty; since we did not vary reward scaling or optimization hyperparameters for this axis, their contribution cannot be ruled out.

This result is consistent with the constrained-RL view that hard requirements should be separated from optimizable preferences~\citep{achiam2017cpo}.
The CER-zone scaffold prevents high perceptual scores from compensating for severe transcript drift, but it also requires the perceptual predictor to provide enough within-zone signal to affect GRPO ranking.
Appendix~\ref{app:arousal_failure} reports the retained training evidence and logging limitations.

\paragraph{Auxiliary cross-language check and other domains.}
Appendix~\ref{app:anime_en} reports an auxiliary English \Anime{} study, summarized in Table~\ref{tab:human_combined}.
It suggests that cross-language use of a style reward can be informative, but the study uses the same Japanese-native listener pool and is not treated as primary evidence for cross-lingual human transfer.
The diagnostics themselves are computed per tuple and are not tied to anime style; extending them to other stylized-speech domains, such as sports commentary or children's content, remains future work.

\paragraph{Practical screening heuristic.}
Before scaling RL with a subjective speech reward, our results suggest four checks:
measure reward-gap calibration with a small A/B study when possible (\S\ref{sec:diagnostic_gap});
measure predictor spread on base-model outputs (\S\ref{sec:diagnostic_spread});
check domain coverage or validate cross-domain use with listeners (Appendix~\ref{app:anime_en});
and inspect whether the predictor has enough within-zone variation relative to constraint penalties (\S\ref{sec:diagnostic_within_zone}).
These checks do not prove causality and are not validated as pre-RL predictors; we propose them to flag rewards likely to produce low-confidence or non-transferable policy movement.

\vspace{-1mm}\section{Conclusion}

We studied when RL from learned subjective predictors transfers from machine-score gains to human-perceptual gains in codec-based speech language models.
For the predictor--axis--base tuples we studied, predictor gains alone are not sufficient: average transfer and within-axis reward-gap calibration can diverge, and both must be considered alongside predictor resolution on the target base distribution, domain validation, and sufficient within-zone signal under intelligibility constraints.
\Anime{} gives a strong in-domain positive case, \Utmos{} shows high machine--human agreement with only modest average preference shift, \Likab{} fails on average but aligns in high-confidence regions, and \Arousal{} fails as a training-time reward under our constrained scaffold.
Consistent with prior machine-level findings, Best-of-$N$ reranking shows that reward-selected samples often already exist in the base model's support. Our listening tests further show that listeners do not clearly prefer GRPO over Best-of-8, indicating that GRPO primarily amortizes reward-based selection into the policy rather than uniformly outperforming reranking. These findings suggest practical screening heuristics for subjective speech rewards before full RL or multi-reward post-training. Overall, all three main predictors can be optimized under the CER gate, while human-perceived transfer depends on the specific tuple; accordingly, the diagnostics in \S\ref{sec:diagnostics} provide a more reliable basis than machine-score gains alone for selecting rewards in multi-reward post-training.

\vspace{-1mm}\section*{Limitations}
\label{sec:limitations}

Our experiments are designed to compare subjective reward behavior under controlled conditions, but they do not fully disentangle reward-model architecture, perceptual axis, and base-model distribution. The reward models differ in training data, score scale, target construct, and base-model coverage; therefore, our results should be read as diagnostics for predictor--axis--base tuples rather than causal claims that one perceptual axis is intrinsically easier or harder than another.

Our evaluation is also limited by practical compute and annotation budgets.
We evaluate one primary backbone and codec, a single main training run per axis, a single primary decoding configuration, a fixed set of reward axes, Best-of-8 as the main reranking baseline, and 50 test items per axis with 5 ratings per item from 10 listeners per axis.
Evaluating an additional speech-LM backbone would be the most informative extension.
Multi-seed training with multi-seed outputs in the human study, more listeners per item, and more diverse prompts would narrow the per-axis confidence intervals; native English listeners for the auxiliary English condition, larger reranking budgets, fixed-KL comparisons, and direct within-prompt rollout-spread measurements would further test the proposed diagnostics.
Within the present budget, we report a residual-violator robustness check (Table~\ref{tab:cer_retry_residual}), Wilson intervals for item-level results (Table~\ref{tab:item_ci}), and item-clustered bootstrap intervals for vote-level results (Table~\ref{tab:vote_ci}), and we treat the pooled vote-level regression over 1,000 votes as the primary calibration evidence because it does not rest on any single per-axis fit.
These intervals and regressions cluster by item but do not model dependence among votes from the same listener, and the JP and EN \Anime{} studies share one listener pool.
These extensions would strengthen the generality of the proposed diagnostics, especially for future multi-reward speech post-training.

\vspace{-1mm}\section*{Ethical Considerations}
\label{sec:ethics}

This work studies post-training methods for controllable synthetic speech.
Such methods can support creative and accessibility-oriented applications, but they can also lower the cost of generating speech in a target style without consent.
We therefore frame our study around evaluation and diagnostics rather than deployment, and we release only artifacts intended for research use: code, prompts, generated audio samples, reward scores, and anonymized human-evaluation votes.
We do not redistribute merged base-model weights, reward model weights, or training corpora whose licenses or copyright status do not permit redistribution.

The \Anime{} reward model targets a stylistic dimension rather than speaker identity, but style and identity can interact in downstream use.
Released materials therefore include responsible-use guidance and are not intended for impersonation, voice cloning, or unauthorized style imitation.
More broadly, subjective speech rewards should be developed with attention to fairness, speaker consent, dataset provenance, and copyright compliance.
Future work should further examine how style-control rewards behave across demographic groups, listener communities, and culturally specific notions of expressiveness or likability.

\paragraph{Use of AI assistance.}
The authors used AI assistants for code writing, language polishing, LaTeX editing assistance, and brainstorming presentation of results. All scientific claims, experiments, analyses, and final manuscript content were verified and revised by the authors.

\bibliography{custom}

\appendix

\vspace{-2mm}\section{Training Hyperparameters and Checkpoint Selection}
\label{app:hparams}

Table~\ref{tab:hparams} reports the common GRPO hyperparameters shared by the reward-axis runs and the per-axis selected checkpoints.
All training is done with \texttt{verl}~\citep{sheng2024verl} on a single H100 80~GB GPU with vLLM-backed rollouts.
The reward predictors run on a second H100 to isolate codec decoding and predictor inference from the rollout engine.

\begin{table*}[t]
\centering
\small
\setlength{\tabcolsep}{4.0pt}
\begin{tabular}{@{}ll@{}}
\toprule
Setting & Value \\
\midrule
Framework / rollout & \texttt{verl} / vLLM (\texttt{gpu\_memory\_utilization}=0.6) \\
Rollouts per prompt $K$ & 4 \\
Train batch size & 16 (PPO mini-batch 16, micro-batch 4 per GPU) \\
Actor optimizer / LR & AdamW / $5\!\times\!10^{-7}$ \\
PPO clip ratio & 0.1 \\
Entropy coefficient & 0 \\
Top-$p$ / Temperature / Rep.\ penalty & 0.85 / 1.0 / 1.05 \\
Max prompt / response length & 512 / 2048 codec tokens \\
Speech-only token mask & enabled \\
KL mode & in-reward (\texttt{use\_kl\_loss=False}) \\
Adaptive KL controller & init $\beta=0.05$, target KL $=0.05$, horizon $=2000$ \\
\texttt{save\_freq} / \texttt{test\_freq} & 30 / 30 steps \\
Hardware & 1$\times$ H100 80~GB for actor + 1$\times$ H100 80~GB for reward predictors \\
Wall time per step & $\sim$115\,s with Whisper in reward, $\sim$43\,s without Whisper \\
\midrule
\multicolumn{2}{c}{\emph{Validation and evaluation sets}} \\
JP validation set & $n=100$ \\
EN validation set & $n=50$ (the EN evaluation prompts) \\
Paper evaluation set & $n=50$ per language \\
\midrule
\multicolumn{2}{c}{\emph{Selected checkpoint step}} \\
\Anime{} JP / EN (Zone-CER) & step 1400 / step 900 \\
\Utmos{} / \Likab{} JP (Zone-CER) & step 1020 / step 540 \\
\Anime{} / \Utmos{} JP (Target-only) & step 540 / step 420 \\
\Arousal{} JP (Zone-CER) & no usable checkpoint \\
\bottomrule
\end{tabular}
\caption{GRPO training hyperparameters.}
\label{tab:hparams}
\end{table*}

\vspace{-2mm}\section{Reward Predictor Implementation}
\label{app:reward_servers}

This appendix documents the reward predictors used in the paper: backbone, checkpoint, training data, output scale, and release plan.
Each predictor receives generated XCodec2 tokens, decodes them to a 16~kHz mono waveform, and returns a scalar score.

\vspace{-2mm}\paragraph{\Anime{}.}
\Anime{} is a pairwise-preference-trained anime-likeness predictor~\citep{park2026animescore}.
It uses a \texttt{microsoft/wavlm-base} self-supervised learning (SSL) encoder with the CNN frozen and the transformer fine-tuned, a learned mixture over the last four hidden layers, a BiLSTM, and a two-layer MLP head.
The model is trained with a RankNet-style pairwise ranking loss on the AnimeScore preference corpus and follows the SSL-MOS fine-tuning recipe of \citet{cooper2022ssl_mos}.
The output is an unbounded scalar; its empirical range on base-model outputs is approximately $[-3,+5]$.
We release the checkpoint with the artifact bundle where licensing permits.

\vspace{-2mm}\paragraph{\Utmos{}.}
\Utmos{} is used off-the-shelf from UTMOS22-strong~\citep{saeki2022utmos}.
No fine-tuning is performed by us.
The output is a naturalness MOS-like score in $[1,5]$, and we divide it by 5 before applying the CER-zone reward.

\vspace{-2mm}\paragraph{\Likab{}.}
\Likab{} is trained for this study on the CocoNut-Humoresque likability dataset~\citep{suda2025coconut_humoresque}.
The model uses \texttt{microsoft/wavlm-base-plus} with the CNN frozen and the transformer fine-tuned, mean pooling, a \texttt{Linear(768,6)} classification head, and a softmax expectation over six discrete classes.
The output is the expected class value on $[1,6]$.
We use the original CocoNut-Humoresque split and train with inverse-frequency class-weighted cross-entropy rather than regression to preserve score spread for GRPO.
Training uses AdamW with learning rate $10^{-5}$ and weight decay $0.01$, linear warmup followed by cosine decay, Gaussian noise augmentation at SNR 30--50~dB, and random gain perturbation of $\pm 3$~dB.
The checkpoint is selected by validation SRCC.
On the Humoresque test set ($n=283$), the model obtains SRCC $0.712$, LCC $0.703$, MSE $0.274$, and prediction range $2.71$.
The classification formulation is intentional: regression counterparts mean-collapse, compressing the prediction range and producing a flatter reward landscape for GRPO.
We release the checkpoint with the artifact bundle, subject to the Humoresque licensing terms.

\vspace{-2mm}\paragraph{\Arousal{}.}
\Arousal{} is the off-the-shelf MSP-Dim valence-arousal-dominance predictor~\citep{wagner2023msp}.
It uses a \texttt{wav2vec2-large-robust} backbone with a regression head emitting valence, arousal, and dominance.
We use only the arousal output and clip it to $[0,1]$.
No fine-tuning is performed by us.

\vspace{-2mm}\paragraph{Whisper / CER.}
For the CER gate, we use Whisper large-v3~\citep{radford2023whisper}.
Language detection is automatic.
CER is computed against the canonical written prompt, with no additional Japanese- or English-specific text normalization.
Thus, minor whitespace and punctuation differences can propagate into CER; this conservative choice is part of the intelligibility signal controlled by the CER-zone reward.

\vspace{-2mm}\paragraph{Reward functions.}
The token-to-reward mapping is implemented separately for each axis: \Anime{}+CER, \Utmos{}+CER, \Likab{}+CER, \Arousal{}+CER, and AS-only.

\vspace{-2mm}\section{Evaluation Set}
\label{app:test_v2}

Table~\ref{tab:test_v2_groups} summarizes the held-out prompt set used for machine evaluation and human listening tests.

\begin{table}[t]
\centering
\small
\begin{tabular}{@{}llr@{}}
\toprule
Group & Description & $n$ \\
\midrule
G1 & emotional expression & 18 \\
G2 & anime-stylized text & 14 \\
G3 & neutral conversational & 10 \\
G4 & long-form narrative & 3 \\
G5 & linguistically challenging & 5 \\
\midrule
Total & & 50 \\
\bottomrule
\end{tabular}
\caption{Composition of test set.}
\label{tab:test_v2_groups}
\end{table}

The test set is held out from reward-model training, GRPO training, and checkpoint-selection validation.
Table~\ref{tab:groups} breaks the target-axis changes of Table~\ref{tab:heatmap4} down by group. Gains are smallest on prompt groups where base scores are already high; \Utmos{} and \Likab{} gain most on anime-stylized texts, where their base scores are lowest, whereas \Anime{} shows the smallest gain on that group. On anime-stylized texts, where base CER is highest (0.31), CER falls by 0.078, 0.088, and 0.099 under the \Anime{}, \Likab{}, and \Utmos{} rewards.

\begin{table}[t]
\centering
\small
\setlength{\tabcolsep}{2.4pt}
\begin{tabular}{@{}lrrrrrrr@{}}
\toprule
& & \multicolumn{2}{c}{\Anime{}} & \multicolumn{2}{c}{\Likab{}} & \multicolumn{2}{c}{\Utmos{}} \\
\cmidrule(lr){3-4}\cmidrule(lr){5-6}\cmidrule(lr){7-8}
Group & $n$ & Base & $\Delta$ & Base & $\Delta$ & Base & $\Delta$ \\
\midrule
Emotional & 18 & $-0.25$ & $+1.42$ & 4.17 & $+0.16$ & 3.00 & $+0.30$ \\
Anime-styl. & 14 & $+0.61$ & $+0.95$ & 3.94 & $+0.38$ & 2.32 & $+1.11$ \\
Neutral & 10 & $-1.29$ & $+1.43$ & 4.46 & $-0.01$ & 3.73 & $+0.20$ \\
Long-form & 3 & $-0.97$ & $+2.24$ & 4.47 & $+0.00$ & 3.79 & $+0.04$ \\
Challeng. & 5 & $-1.56$ & $+1.56$ & 4.42 & $+0.04$ & 3.77 & $+0.23$ \\
\bottomrule
\end{tabular}
\caption{Target-predictor score of base outputs and its change under the matching single-reward GRPO policy, by prompt group (first-shot outputs, as in Table~\ref{tab:heatmap4}).}
\label{tab:groups}
\end{table}

\vspace{-2mm}\paragraph{Residual violations after retry.}
Table~\ref{tab:cer_retry_residual} reports how often either side remains above $\tau_h=0.30$ after the CER-retry pass.
The union column gives the number of item pairs excluded in the clean-only robustness check.
Here the target side is the GRPO output in the GRPO-vs-base comparison.

\begin{table}[t]
\centering
\small
\setlength{\tabcolsep}{3.0pt}
\begin{tabular}{@{}lrrr@{}}
\toprule
Axis & base resid. & GRPO resid. & union resid. \\
\midrule
\Anime{} JP & $4/50$ & $3/50$ & $4/50$ \\
\Anime{} EN & $1/50$ & $1/50$ & $1/50$ \\
\Utmos{} & $3/50$ & $3/50$ & $4/50$ \\
\Likab{} & $6/50$ & $3/50$ & $7/50$ \\
\bottomrule
\end{tabular}
\caption{Residual CER $>0.30$ after the CER-retry pass.
The union residual count is the number of item pairs where either side remains above the CER threshold.}
\label{tab:cer_retry_residual}
\end{table}

\vspace{-2mm}\paragraph{Vote-level robustness.}
Table~\ref{tab:cer_retry_robustness_vote} reports vote-level Human-WR and agreement before and after excluding item pairs where either side remains above $\tau_h$ after retry.
The ``all'' rows match the vote-level summaries in Appendix~\ref{app:human_stats}.

\begin{table*}[t]
\centering
\small
\setlength{\tabcolsep}{3.0pt}
\begin{tabular}{@{}lrrrrrrr@{}}
\toprule
Axis & excl. pairs & HWR all & HWR clean & $\Delta$HWR & Agree all & Agree clean & $\Delta$Agr \\
\midrule
\Anime{} JP & $4$ & $72.0$ & $73.9$ & $+1.9$ pp & $78.4$ & $80.9$ & $+2.5$ pp \\
\Anime{} EN & $1$ & $62.8$ & $62.9$ & $+0.1$ pp & $68.8$ & $69.4$ & $+0.6$ pp \\
\Utmos{} & $4$ & $56.4$ & $54.3$ & $-2.1$ pp & $72.0$ & $71.3$ & $-0.7$ pp \\
\Likab{} & $7$ & $38.0$ & $33.5$ & $-4.5$ pp & $67.6$ & $66.5$ & $-1.1$ pp \\
\bottomrule
\end{tabular}
\caption{Vote-level Human-WR and machine--human agreement before and after excluding residual CER-violator item pairs.
Clean-only percentages are computed after removing all five votes for each excluded item pair.
The clean-only analysis changes the exact percentages but does not change the qualitative conclusions: \Anime{} remains positive, \Utmos{} remains modestly positive, and \Likab{} remains the negative average-transfer case.}
\label{tab:cer_retry_robustness_vote}
\end{table*}

\vspace{-2mm}\paragraph{Logistic check.}
The reward-gap regression in Appendix~\ref{app:gap_regression} further shows that the pooled CER-gap slope is statistically indistinguishable from zero, while the reward-gap slope is significant.
Together with Table~\ref{tab:cer_retry_robustness_vote}, this indicates that residual CER differences are unlikely to be the primary driver of the main human-evaluation conclusions.

\vspace{-2mm}\section{Human Recruitment}
\label{app:artifacts}

\vspace{-2mm}\paragraph{Collecting human-evaluation data.}
Human-evaluation votes are released with randomized item identifiers and anonymized listener identifiers.
We do not release platform-specific worker identifiers or personally identifying information.
We collected basic self-reported metadata, including demographics, for quality control and aggregate analysis.

\begin{figure*}[t]
\centering
\begin{minipage}{0.48\linewidth}
    \centering
    \includegraphics[width=\linewidth]{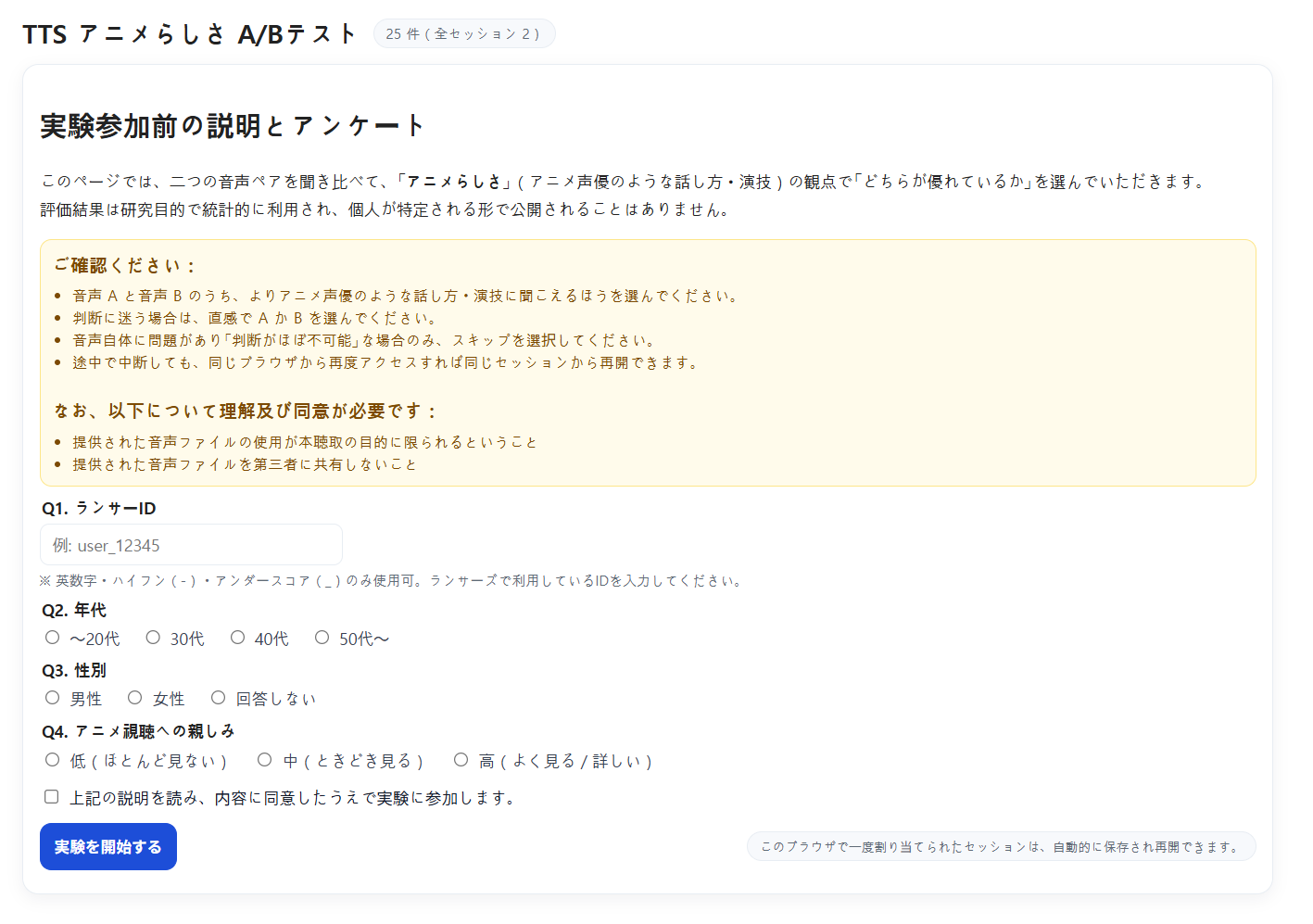}
    \vspace{-2mm}
    \small (a) Pre-task instruction and consent page.
\end{minipage}
\hfill
\begin{minipage}{0.48\linewidth}
    \centering
    \includegraphics[width=\linewidth]{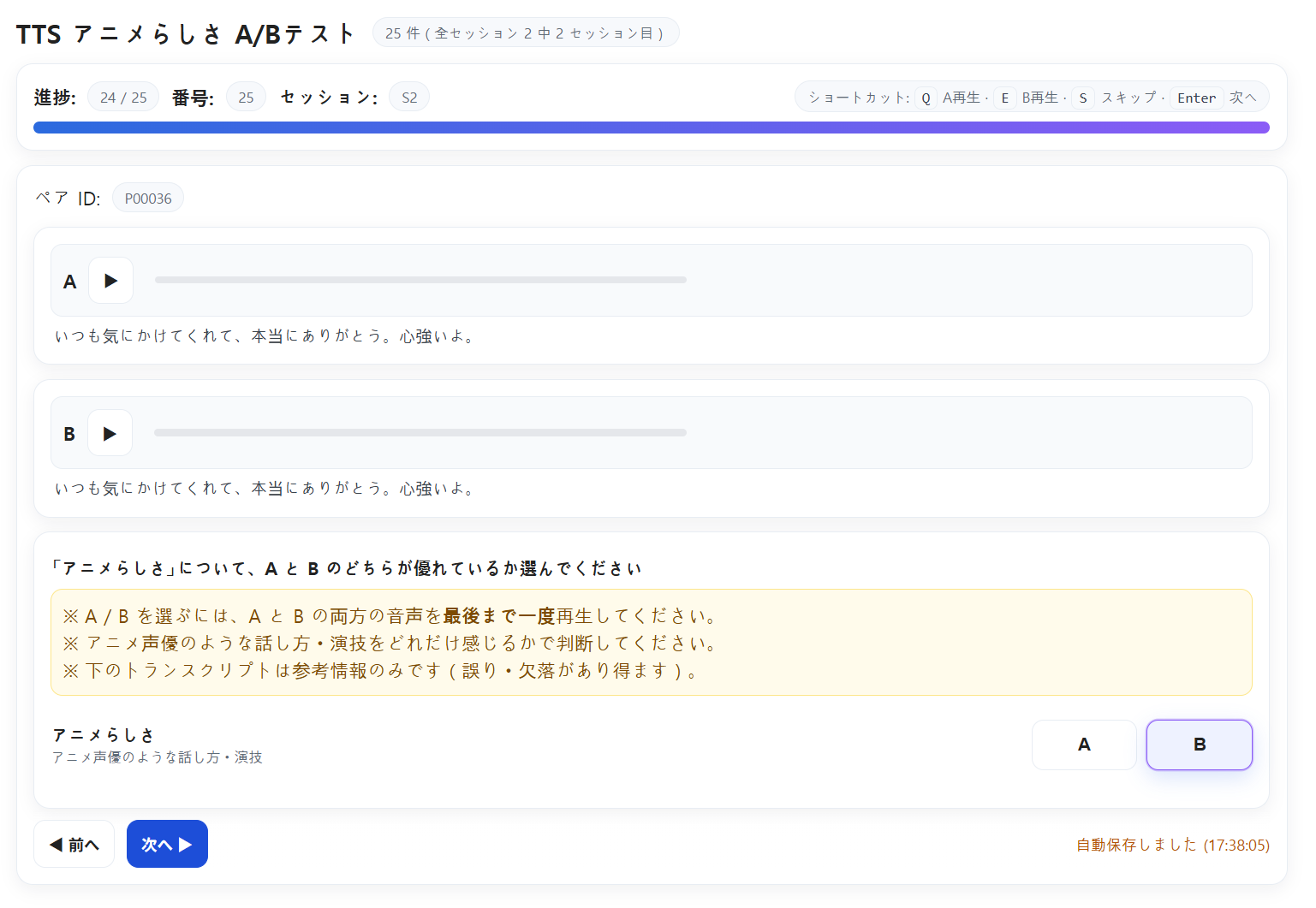}
    \vspace{-2mm}
    \small (b) Pairwise A/B listening interface.
\end{minipage}
\vspace{-2mm}
\caption{Screenshots of the human-evaluation interface.
Before starting, participants read the task description, consent notice, and demographic questions.
During evaluation, they listened to both clips in each pair and selected the side that better matched the target perceptual criterion.}
\label{fig:human_eval_ui}
\end{figure*}

\vspace{-2mm}\paragraph{Participant recruitment and compensation.}
Listeners were recruited through Lancers, a Japanese crowdsourcing platform.
Each listening session contained 25 A/B pairs, took approximately 10 minutes, and was compensated at 200 JPY per completed session.
Before starting, workers were shown task instructions explaining the target comparison question, playback requirement, skip conditions, and research use of the responses.
The human-evaluation interface included a pre-task instruction and consent page and a pairwise A/B listening page, shown in Figure~\ref{fig:human_eval_ui}.

\vspace{-2mm}\section{Human Evaluation Statistics}
\label{app:human_stats}

This appendix provides confidence intervals and vote-level summaries for the human-evaluation results.
Table~\ref{tab:item_ci} reports item-level Wilson intervals over the 50 majority-vote items for GRPO vs Base.
Table~\ref{tab:bon_item_ci} reports the corresponding item-level intervals for GRPO vs Best-of-8.
Table~\ref{tab:vote_ci} reports vote-level summaries with item-clustered bootstrap intervals.
The English \Anime{} row is auxiliary and is discussed separately in Appendix~\ref{app:anime_en}.

\begin{table*}[t]
\centering
\small
\setlength{\tabcolsep}{3.0pt}
\begin{tabular}{@{}lrrrrrr@{}}
\toprule
Axis & HWR & HWR CI & MWR & MWR CI & Agree & Agree CI \\
\midrule
\Anime{} JP & 80.0 & [67.0,88.8] & 88.0 & [76.2,94.4] & 88.0 & [76.2,94.4] \\
\Utmos{} & 62.0 & [48.2,74.1] & 74.0 & [60.4,84.1] & 80.0 & [67.0,88.8] \\
\Likab{} & 36.0 & [24.1,49.9] & 56.0 & [42.3,68.8] & 76.0 & [62.6,85.7] \\
\Anime{} EN & 70.0 & [56.2,80.9] & 66.0 & [52.2,77.6] & 72.0 & [58.3,82.5] \\
\bottomrule
\end{tabular}
\caption{Item-level Wilson 95\% confidence intervals over 50 items for GRPO vs Base.
HWR, MWR, and agreement are item-level percentages computed from 5-rater majority votes.}
\label{tab:item_ci}
\end{table*}

\begin{table*}[t]
\centering
\small
\setlength{\tabcolsep}{3.0pt}
\begin{tabular}{@{}lrrrrrr@{}}
\toprule
Axis & HWR & HWR CI & MWR & MWR CI & Agree & Agree CI \\
\midrule
\Anime{} & 52.0 & [38.5,65.2] & 50.0 & [36.6,63.4] & 74.0 & [60.4,84.1] \\
\Utmos{} & 46.0 & [33.0,59.6] & 36.0 & [24.1,49.9] & 68.0 & [54.2,79.2] \\
\Likab{} & 48.0 & [34.8,61.5] & 32.0 & [20.8,45.8] & 62.0 & [48.2,74.1] \\
\bottomrule
\end{tabular}
\caption{Item-level Wilson 95\% confidence intervals for GRPO vs Best-of-8 over 50 paired items.
HWR, MWR, and agreement are item-level percentages computed from 5-rater majority votes.}
\label{tab:bon_item_ci}
\end{table*}

\begin{table}[t]
\centering
\small
\setlength{\tabcolsep}{3.0pt}
\begin{tabular}{@{}lrrrr@{}}
\toprule
Term & $\hat\beta$ & SE & $z$ & $p$ \\
\midrule
Intercept & {$+0.868$} & {$0.145$} & {$+5.99$} & {$2.1\!\times\!10^{-9}$} \\
\Anime{} JP & {$+0.553$} & {$0.224$} & {$+2.47$} & {$0.014$} \\
\Likab{} & {$-1.235$} & {$0.202$} & {$-6.11$} & {$1.0\!\times\!10^{-9}$} \\
\Utmos{} & {$-0.362$} & {$0.210$} & {$-1.73$} & {$0.084$} \\
$z_R$ & {$\mathbf{+1.660}$} & {$0.132$} & {$+12.57$} & {$\mathbf{3.3\!\times\!10^{-36}}$} \\
$z_C$ & {$+0.142$} & {$0.100$} & {$+1.42$} & {$0.157$} \\
\bottomrule
\end{tabular}
\caption{Logistic regression of human pair-preference on standardized reward gap and CER gap.
$\mathrm{OR}(z_R)={5.26}$ with 95\% CI {$[4.06,6.81]$}, while $\mathrm{OR}(z_C)={1.15}$ with 95\% CI {$[0.95,1.40]$}.}
\label{tab:logit_main}
\end{table}

\begin{table}[t]
\centering
\small
\setlength{\tabcolsep}{3.0pt}
\begin{tabular}{@{}lrrr@{}}
\toprule
Metric & base & GRPO & $\Delta$ \\
\midrule
\Anime{} mean & $-0.55$ & {$-0.05$} & {$+0.50$} \\
CER mean & $0.044$ & $0.034$ & $-0.010$ \\
CER median & {$0.014$} & {$0.009$} & {$-0.005$} \\
\Utmos{} & $3.10$ & $3.18$ & $+0.08$ \\
\bottomrule
\end{tabular}
\caption{\Anime{} EN base vs.\ Zone-CER step 900 GRPO on English test set, $n=50$.}
\label{tab:anime_en_objective}
\end{table}

\begin{table}[t]
\centering
\small
\setlength{\tabcolsep}{3.0pt}
\begin{tabular}{@{}lr@{}}
\toprule
Quantity & Value \\
\midrule
Total training steps & 1730 \\
Validation evaluations & 60 \\
Init $\beta$ / final $\beta$ & 0.050 / 0.082 \\
$\beta$ range & $[0.002,0.122]$ \\
Plateau $\beta$ median & 0.025 \\
Plateau $\beta$ IQR & $[0.014,0.052]$ \\
Plateau reward-KL penalty median & 0.041 \\
Max reward-KL penalty & 0.094 \\
\bottomrule
\end{tabular}
\caption{Adaptive-KL statistics for the primary \Anime{} JP Zone-CER run.
Plateau is defined as steps $\ge 300$.}
\label{tab:kl_traj}
\end{table}

\begin{table}[t]
\centering
\small
\setlength{\tabcolsep}{3.0pt}
\begin{tabular}{@{}rrrrr@{}}
\toprule
Step & val\_AS & val\_CER & val\_viol. & val\_reward \\
\midrule
1050 & $-0.138$ & $0.230$ & $0.27$ & $1.902$ \\
1380 & $+0.562$ & $0.244$ & $0.24$ & $2.579$ \\
\textbf{1400} & {\boldmath$+0.625$} & {\boldmath$0.268$} & {\boldmath$0.26$} & {\boldmath$2.584$} \\
1710 & $+0.955$ & $0.308$ & $0.31$ & $2.556$ \\
\bottomrule
\end{tabular}
\caption{Validation trajectory at key steps for the primary \Anime{} JP Zone-CER run.
Raw \Anime{} continues rising after step 1400, but constraint-aware validation reward declines as violations increase.}
\label{tab:val_traj}
\end{table}

\begin{table*}[t]
\centering
\small
\setlength{\tabcolsep}{4.0pt}
\begin{tabular}{@{}lrrrr@{}}
\toprule
Axis & GRPO votes & Vote HWR & Vote Agree & Cluster-boot CI \\
\midrule
\Anime{} JP & 180/250 & 72.0 & 78.4 & [64.0,79.6] / [72.4,84.4] \\
\Utmos{} & 141/250 & 56.4 & 72.0 & [47.2,65.2] / [64.8,78.8] \\
\Likab{} & 95/250 & 38.0 & 67.6 & [29.2,47.2] / [58.8,76.0] \\
\Anime{} EN & 157/250 & 62.8 & 68.8 & [54.8,70.4] / [62.4,75.2] \\
\bottomrule
\end{tabular}
\caption{Vote-level summaries for GRPO vs Base with item-clustered bootstrap 95\% confidence intervals.
The two intervals in the last column correspond to Vote HWR and vote-level agreement.}
\label{tab:vote_ci}
\end{table*}

\vspace{-2mm}\section{Reward-Gap Logistic Regression}
\label{app:gap_regression}

\vspace{-2mm}\paragraph{Outcome variable.}
Let $y_{ij}\in\{0,1\}$ be listener $i$'s preference on item $j$.
We set $y_{ij}=1$ if the listener chooses the target side, i.e., the GRPO output in the GRPO-vs-base comparisons.
The signed reward gap $\Delta r=r_{\mathrm{target}}-r_{\mathrm{base}}$ indicates how strongly the target reward favors the GRPO side.

\vspace{-2mm}\paragraph{Predictors.}
$z_R$ is the within-axis $z$-score of $\Delta r$.
$z_C$ is the within-axis $z$-score of $\Delta \mathrm{CER}=\mathrm{CER}_{\mathrm{base}}-\mathrm{CER}_{\mathrm{target}}$, so positive $z_C$ indicates that the target side is more intelligible.

\vspace{-2mm}\paragraph{Data.}
The regression uses 1000 vote-level observations from 200 item pairs across four axes: \Anime{} EN, \Anime{} JP, \Likab{}, and \Utmos{}.
Each axis contains 50 item pairs with 5 listener votes per item.
Rows produced only for visualization are not used in this regression.

\vspace{-2mm}\paragraph{Primary fit.}
We fit a logistic regression with axis fixed effects and item-clustered robust standard errors.
Table~\ref{tab:logit_main} reports the primary pooled fit.
The reference axis is \Anime{} EN.

\vspace{-2mm}\paragraph{Equality test and likelihood-ratio tests.}
A Wald test rejects equality of slopes: $\beta_R-\beta_C={+1.518}$, $\mathrm{SE}={0.164}$, $z={+9.26}$, and $p={2.1\!\times\!10^{-20}}$.
Adding $z_R$ to a model with axis fixed effects and $z_C$ improves fit by $\chi^2_1={301.45}$ with $p={1.6\!\times\!10^{-67}}$.
Adding $z_C$ to a model with axis fixed effects and $z_R$ does not improve fit, with $\chi^2_1={2.68}$ and $p={0.10}$.

\vspace{-2mm}\paragraph{Per-axis fits.}
Table~\ref{tab:logit_peraxis} reports descriptive per-axis logistic fits.

\begin{table*}[t]
\centering
\small
\setlength{\tabcolsep}{3.0pt}
\begin{tabular}{@{}lrrr@{}}
\toprule
Axis & $\hat\beta_R$ (SE, $p$) & $\hat\beta_C$ (SE, $p$) & OR$_R$ \\
\midrule
\Anime{} EN & {$+1.43$ ($0.24$, $<0.001$)} & {$+0.13$ ($0.19$, $0.48$)} & {$4.17$} \\
\Anime{} JP & {$+1.40$ ($0.26$, $<0.001$)} & {$+0.15$ ($0.22$, $0.50$)} & {$4.05$} \\
\Likab{} & {$+2.11$ ($0.34$, $<0.001$)} & {$+0.03$ ($0.19$, $0.88$)} & {$8.27$} \\
\Utmos{} & {$+1.95$ ($0.22$, $<0.001$)} & {$+0.18$ ($0.20$, $0.39$)} & {$7.03$} \\
\bottomrule
\end{tabular}
\caption{Per-axis logistic fits.
$\hat\beta_R$ is positive with $p<0.001$ in every axis, and $\hat\beta_C$ has $p>0.3$ in every axis.}
\label{tab:logit_peraxis}
\end{table*}

\vspace{-2mm}\paragraph{Multicollinearity.}
The pooled Pearson correlation between $z_R$ and $z_C$ is $r=+0.16$.
Within axis, the correlations are $+0.16$ for \Anime{} EN, $+0.13$ for \Anime{} JP, $+0.40$ for \Likab{}, and $-0.05$ for \Utmos{}.
The two predictors are therefore weakly correlated, so the joint-model coefficients are interpretable as partial effects.

\vspace{-2mm}\paragraph{Sensitivity to EN inclusion.}
Restricting the regression to the three JP axes preserves the qualitative pattern that reward-gap slopes are positive.
We report the four-axis pooled regression in the main text because it has the broadest coverage and because the EN row is part of the auxiliary evidence for cross-language reward transfer.
The main conclusion remains that standardized reward gaps predict human preference more strongly than residual CER gaps.

\vspace{-2mm}\section{Reward-Gap Binned Analysis}
\label{app:gap_bins}

This appendix gives a descriptive reward-gap bin analysis underlying the main-text calibration discussion.
The formal test is the signed-gap logistic regression in Appendix~\ref{app:gap_regression}; the binned analysis is intended only to visualize low- and high-gap regimes.
Because the bins are based on absolute reward-gap magnitude, the agreement column is the more direct direction-invariant summary, while the GRPO preference column shows how often listeners choose the optimized output within each bin.
Table~\ref{tab:gap_bins} reports the resulting vote-level bins.

\begin{table*}[t]
\centering
\small
\setlength{\tabcolsep}{3.0pt}
\begin{tabular}{@{}llrrrr@{}}
\toprule
Axis & $|\Delta\mathrm{reward}|$ bin & $n$ items & $n$ votes & Vote HWR & Vote Agree \\
\midrule
\Anime{} JP & 0--0.3 & 5 & 25 & 28.0 & 64.0 \\
\Anime{} JP & 0.3--1.0 & 8 & 40 & 62.5 & 62.5 \\
\Anime{} JP & 1.0--2.5 & 19 & 95 & 65.3 & 72.6 \\
\Anime{} JP & 2.5--$\infty$ & 18 & 90 & 95.6 & 95.6 \\
\midrule
\Anime{} EN & 0--0.3 & 17 & 85 & 48.2 & 49.4 \\
\Anime{} EN & 0.3--1.0 & 18 & 90 & 61.1 & 71.1 \\
\Anime{} EN & 1.0--2.5 & 11 & 55 & 74.5 & 83.6 \\
\Anime{} EN & 2.5--$\infty$ & 4 & 20 & 100.0 & 100.0 \\
\midrule
\Utmos{} & 0--0.3 & 17 & 85 & 45.9 & 52.9 \\
\Utmos{} & 0.3--1.0 & 24 & 120 & 48.3 & 75.8 \\
\Utmos{} & 1.0--2.5 & 9 & 45 & 97.8 & 97.8 \\
\midrule
\Likab{} & 0--0.3 & 40 & 200 & 24.5 & 61.5 \\
\Likab{} & 0.3--1.0 & 8 & 40 & 90.0 & 90.0 \\
\Likab{} & 1.0--2.5 & 2 & 10 & 100.0 & 100.0 \\
\bottomrule
\end{tabular}
\caption{Descriptive vote-level reward-gap binned analysis.
Items are grouped by absolute target-reward gap.
Vote HWR is the fraction of individual votes choosing GRPO, and Vote Agree is the fraction of individual votes matching the machine-preferred side.
This table complements the signed-gap regression in Appendix~\ref{app:gap_regression}.
High-gap \Likab{} bins contain few item pairs and should not be interpreted as standalone statistical evidence.}
\label{tab:gap_bins}
\end{table*}

\vspace{-2mm}\section{English AnimeScore Evaluation}
\label{app:anime_en}

We repeat the JP \Anime{} experiment in English as a separate GRPO run: using the same \Anime{} predictor, we train on 900 English prompts and evaluate on a 50-prompt English test set.
The EN test prompts are idiomatic English counterparts of the JP test items with the same five-group structure, written for English listeners rather than translated literally.
Both the EN training set and the EN test set are disjoint from the JP corpora.
Checkpoint selection for this run (step 900) used these 50 English prompts.

\vspace{-2mm}\paragraph{Objective results.}
On the EN test set with the same CER-retry protocol, Zone-CER GRPO at step 900 yields $\Delta\Anime{}=+0.50$ over base. Table~\ref{tab:anime_en_objective} summarizes the corresponding objective scores.

\vspace{-2mm}\paragraph{Human results.}
On the same 50 EN items, item-level Human-WR is $70.0\%$ with Wilson CI $[56.2,80.9]$.
Machine-WR is $66.0\%$, item-level machine--human agreement is $72.0\%$, vote-level Human-WR is $62.8\%$, and vote-level agreement is $68.8\%$.

\vspace{-2mm}\paragraph{Limitation.}
The EN study was rated by the same Japanese-native listener pool that produced the JP results.
It therefore tests whether a JP-trained anime-likeness reward transfers to EN audio under the same listener community, not whether English-native listeners would judge the shift similarly.
We treat the EN result as auxiliary evidence, not as primary evidence for cross-lingual human transfer.

\vspace{-2mm}\section{VAD-Arousal Training-Only Negative Result}
\label{app:arousal_failure}

This appendix documents the training-only negative result for \Arousal{}.
No human A/B study was conducted for this axis, so the evidence concerns training dynamics rather than perceptual transfer.

Validation arousal stayed in the range $0.61$--$0.64$ during early training, yielding a net change of only $\Delta\approx+0.014$.
At the same time, validation reward improved from $-0.34$ to $-0.05$, while the validation violation rate dropped from $0.62$ to $0.45$.
This suggests that training primarily improved reward by moving samples out of the VIOLATE shelf, rather than by increasing arousal within the feasible zone.

Per-rollout reward decompositions were not retained, so this result is treated as a training-dynamics diagnostic rather than a causal attribution of individual GRPO updates.

\section{Adaptive-KL Statistics}
\label{app:kl_csv}

Table~\ref{tab:kl_traj} summarizes the adaptive-KL controller for the primary \Anime{} JP Zone-CER run. Table~\ref{tab:val_traj} reports key validation points used for checkpoint selection.

\vspace{-2mm}\paragraph{Step selection.}
Between steps 1400 and 1710, raw validation \Anime{} rises from $+0.625$ to $+0.955$, but the validation violation rate also rises from $0.26$ to $0.31$ and constraint-aware validation reward decreases from $2.584$ to $2.556$.
We therefore select step 1400.

\input{appendix_qualitative}

\end{document}

%% file: appendix_qualitative.tex
\vspace{-2mm}\section{Qualitative Comparison of GRPO and Best-of-8}
\label{app:qual_bon}

This appendix complements \S\ref{sec:human_bon} by describing how the GRPO and Best-of-8 outputs compared there differ in their acoustic and prosodic properties.
For each axis we use the 50 GRPO clips and the 50 Best-of-8 clips, with the base clips as a reference, as released with the paper and playable on the demo page.
We describe them using their predictor scores and CER from the CER-retry evaluation reported in Table~\ref{tab:human_combined}, their Whisper transcripts, and the acoustic measurements below; descriptions of individual clips are based on these measurements and transcripts.

\vspace{-2mm}\paragraph{Measurements.}
For each clip we use Praat through Parselmouth~\citep{jadoul2018parselmouth} to measure the following.
\begin{itemize}
    \item Speech rate: morae per second over the span between the first and last frames within 30~dB of the clip's peak intensity. Mora counts come from the OpenJTalk reading of the prompt.
    \item Articulation rate: the same rate with internal pauses of at least 150~ms excluded, and the pause ratio, the proportion of the span spent in such pauses.
    \item Median fundamental frequency (F0) and the F0 range between the 5th and 95th percentiles, in semitones.
    \item The standard deviation of frame intensity, and the harmonics-to-noise ratio (HNR) as a coarse indicator of breathiness and noise.
\end{itemize}
All systems are measured on the same 16~kHz MP3 release files.
On every example clip in Table~\ref{tab:qual_examples}, a second pitch tracker, Harvest~\citep{morise2017harvest}, agrees with Praat on median F0 within 2.1 semitones.
Differences are paired by item, with 95\% bootstrap confidence intervals over the 50 items (Table~\ref{tab:qual_acoustic}).

\begin{table*}[t]
\centering
\small
\setlength{\tabcolsep}{3.4pt}
\begin{tabular}{@{}lrrrrrrrrrrrr@{}}
\toprule
& \multicolumn{4}{c}{\Anime{}} & \multicolumn{4}{c}{\Utmos{}} & \multicolumn{4}{c}{\Likab{}} \\
\cmidrule(lr){2-5}\cmidrule(lr){6-9}\cmidrule(lr){10-13}
Measure & Base & G$-$B & B8$-$B & G$-$B8 & Base & G$-$B & B8$-$B & G$-$B8 & Base & G$-$B & B8$-$B & G$-$B8 \\
\midrule
Speech rate (morae/s) & 5.86 & \textbf{\boldmath$+$0.53} & $-$0.12 & \textbf{\boldmath$+$0.65} & 5.87 & \textbf{\boldmath$+$0.45} & $-$0.11 & \textbf{\boldmath$+$0.55} & 5.87 & \textbf{\boldmath$+$0.39} & $+$0.15 & \textbf{\boldmath$+$0.24} \\
Articulation rate (morae/s) & 6.88 & \textbf{\boldmath$+$0.68} & $+$0.12 & \textbf{\boldmath$+$0.57} & 6.91 & \textbf{\boldmath$+$0.43} & $+$0.01 & \textbf{\boldmath$+$0.42} & 6.91 & \textbf{\boldmath$+$0.43} & $+$0.24 & $+$0.18 \\
Pause ratio (\%) & 15.3 & $+$0.2 & \textbf{\boldmath$+$3.7} & \textbf{\boldmath$-$3.5} & 16.2 & $-$2.3 & $+$1.5 & \textbf{\boldmath$-$3.8} & 16.2 & $-$1.1 & $+$0.7 & $-$1.7 \\
Median F0 (st) & 16.5 & $+$0.6 & \textbf{\boldmath$-$2.2} & \textbf{\boldmath$+$2.9} & 16.4 & \textbf{\boldmath$-$3.8} & \textbf{\boldmath$-$2.7} & $-$1.1 & 16.4 & $+$0.2 & \textbf{\boldmath$-$1.3} & \textbf{\boldmath$+$1.4} \\
F0 range (st) & 11.6 & 0.0 & $+$0.9 & $-$0.9 & 11.7 & $-$0.6 & $+$0.4 & \textbf{\boldmath$-$1.0} & 11.7 & $-$0.5 & $+$0.5 & \textbf{\boldmath$-$1.0} \\
Intensity SD (dB) & 6.96 & \textbf{\boldmath$-$0.23} & $-$0.14 & $-$0.08 & 6.99 & \textbf{\boldmath$-$0.17} & $-$0.10 & $-$0.07 & 6.99 & \textbf{\boldmath$-$0.28} & $-$0.07 & \textbf{\boldmath$-$0.21} \\
HNR (dB) & 15.2 & \textbf{\boldmath$-$0.7} & \textbf{\boldmath$-$1.1} & $+$0.4 & 15.4 & $-$0.9 & $-$0.4 & $-$0.5 & 15.4 & 0.0 & $-$0.1 & $+$0.1 \\
\bottomrule
\end{tabular}
\caption{Acoustic comparison of base (B), GRPO (G), and Best-of-8 (B8) clips on the 50 test items of each axis.
Base is the mean over the base clips; the other columns are mean paired differences, with pause-ratio differences in percentage points.
Median F0 is in semitones relative to 100~Hz.
Bold: the paired bootstrap 95\% confidence interval over items excludes zero.}
\label{tab:qual_acoustic}
\end{table*}

\begin{table*}[t]
\centering
\footnotesize
\setlength{\tabcolsep}{3.6pt}
\begin{tabular}{@{}llp{4.9cm}ccccc@{}}
\toprule
Axis & Item & Line (English gloss) & $R$ & CER & Rate & F0 (Hz) & Range (st) \\
\midrule
\multirow{3}{*}{\Anime{}} & S36 & ``Thank you so much for always looking out for me. It really reassures me.'' & $+$3.05 / $-$0.11 & 0.12 / 0.00 & 5.9 / 6.2 & 322 / 278 & 13.1 / 12.5 \\
 & S34 & ``Gazing absently at the view outside the window calms me down.'' & $+$0.26 / $+$3.73 & 0.03 / 0.27 & 7.0 / 6.7 & 223 / 190 & 9.0 / 6.3 \\
 & S32 & ``No way! I can't believe this is actually happening!'' & $+$3.64 / $+$3.62 & 0.14 / 0.29 & 5.0 / 3.5 & 214 / 485 & 12.5 / 13.5 \\
\midrule
\multirow{3}{*}{\Utmos{}} & S16 & ``Good morning! Let's play together again today, promise!'' & 3.23 / 1.52 & 0.27 / 0.27 & 6.1 / 3.0 & 197 / 393 & 7.6 / 13.3 \\
 & S03 & ``I can't take it anymore! Why won't anyone listen to me?'' & 1.66 / 3.26 & 0.19 / 0.19 & 3.5 / 6.0 & 316 / 255 & 3.3 / 8.6 \\
 & S27 & ``Every morning I check the news over coffee.'' & 4.32 / 4.33 & 0.00 / 0.00 & 7.8 / 7.5 & 119 / 149 & 11.5 / 15.6 \\
\midrule
\multirow{3}{*}{\Likab{}} & S32 & ``No way! I can't believe this is actually happening!'' & 4.47 / 3.88 & 0.14 / 0.18 & 6.1 / 5.8 & 287 / 402 & 11.0 / 13.4 \\
 & S04 & ``Everyone, gather round! You can't miss today's event!'' & 3.81 / 4.21 & 0.12 / 0.08 & 5.4 / 5.6 & 469 / 191 & 8.2 / 6.9 \\
 & S34 & ``Gazing absently at the view outside the window calms me down.'' & 4.57 / 4.57 & 0.13 / 0.10 & 6.8 / 6.7 & 257 / 253 & 10.2 / 9.7 \\
\bottomrule
\end{tabular}
\caption{Representative GRPO (G) and Best-of-8 (B8) pairs, three per axis: the largest GRPO advantage in target score, the largest Best-of-8 advantage, and the smallest gap, among the 45 pairs per axis in which both clips have CER $\le 0.30$.
Each cell gives G / B8. $R$ is the raw target-predictor score, rate is in morae/s, F0 is the median, and range is the 5th--95th percentile F0 range.
Item IDs index the released Japanese test prompts.}
\label{tab:qual_examples}
\end{table*}

\begin{figure*}[t]
\centering
\includegraphics[width=\textwidth]{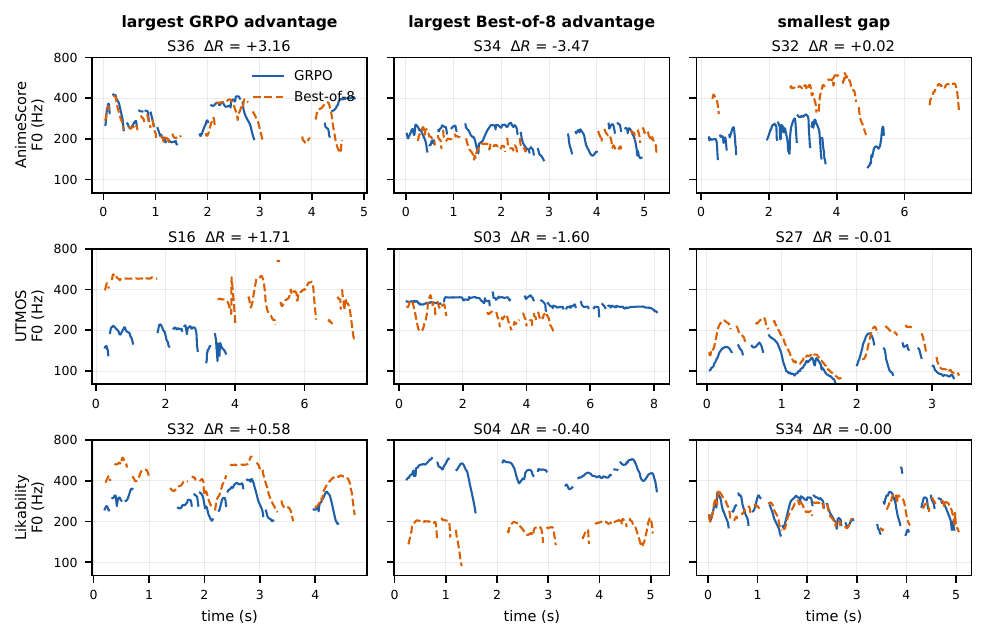}
\caption{F0 contours of the pairs in Table~\ref{tab:qual_examples} (GRPO solid, Best-of-8 dashed; log-frequency axis).
$\Delta R$ is the GRPO minus Best-of-8 target-predictor score.}
\label{fig:qual_f0}
\end{figure*}

\vspace{-2mm}\paragraph{Systematic differences.}
Best-of-8 can only select from base samples, and its speech rate remains close to the base level on all three axes, with mean changes ranging from $-0.12$ to $+0.15$ morae/s and all intervals including zero.
In contrast, all three GRPO policies produce faster speech than the base by $0.39$ to $0.53$ morae/s (7--9\%), and faster than Best-of-8 by $0.24$ to $0.65$ morae/s.
On \Anime{} and \Utmos{}, GRPO clips also pause less than the Best-of-8 outputs.
Because this shift appears under all three rewards, it is not specific to one predictor.

The two methods also differ in pitch characteristics.
Best-of-8 selects lower-pitched outputs than the base on every axis, by 1.3 to 2.7 semitones in median F0, whereas GRPO does so only under \Utmos{} (3.8 semitones lower).
Under \Anime{} and \Likab{}, GRPO clips are therefore 2.9 and 1.4 semitones higher than the Best-of-8 outputs.
Under \Utmos{} and \Likab{}, GRPO also has an F0 range about 1 semitone narrower than Best-of-8, indicating slightly flatter pitch contours.
The spread of voices across items, measured as the between-item standard deviation of median F0, does not differ reliably between GRPO and Best-of-8 on any axis, providing no evidence of voice collapse under GRPO.
HNR differences between GRPO and Best-of-8 are small on every axis, and all intervals include zero.
Under \Anime{}, both are 0.7--1.1~dB below the base.

\vspace{-2mm}\paragraph{Examples.}
Table~\ref{tab:qual_examples} and Figure~\ref{fig:qual_f0} show three pairs per axis, chosen by a fixed rule on the target-score gap $\Delta R$ (GRPO minus Best-of-8), illustrating three recurring patterns.

First, equal scores can arise from very different renditions.
For S32 under \Anime{} ($\Delta R=+0.02$), GRPO reads the line in a mid-range voice at 5.0 morae/s.
The Best-of-8 clip uses a much higher voice (median F0 485 vs.\ 214~Hz) at 3.5 morae/s, and its Whisper transcript renders the opening exclamation as a sigh.
Under \Likab{}, the S04 clips occupy different pitch registers (469 vs.\ 191~Hz), while their scores differ by only 0.4.

Second, the largest gaps occur when the predictor assigns a low score to one of the two outputs.
Under \Utmos{}, the GRPO clip for S03 delivers an exasperated line slowly (3.5 morae/s) on an almost flat pitch (a range of 3.3 semitones), scoring 1.6 below Best-of-8.
Conversely, for S16, the Best-of-8 clip stretches the opening greeting into a long, high, held tone (3.0 morae/s), scoring 1.7 below GRPO.
Under \Anime{}, the higher-pitched voice receives the higher score in S36 (322 vs.\ 278~Hz, $\Delta R=+3.16$).
In S34, the Best-of-8 clip scores higher despite being lower, slower, and narrower in range than the GRPO clip (190 vs.\ 223~Hz, 6.3 vs.\ 9.0 semitones, $\Delta R=-3.47$).

Third, the two methods sometimes produce nearly identical realizations.
For S34 under \Likab{} ($\Delta R=0.00$), the two clips match in tempo (6.8 vs.\ 6.7 morae/s), register (257 vs.\ 253~Hz), and pitch range (10.2 vs.\ 9.7 semitones), and their contours nearly coincide even though they come from different sampling seeds.

Transcripts differ from the target text on both sides; from transcripts alone we cannot distinguish ASR errors from pronunciation errors.
In S36, the Whisper transcript of the GRPO clip differs in the final word (\emph{kokorozuyoi}, ``reassuring'').
In S34 under \Anime{}, the transcript of the Best-of-8 clip differs in two content words (CER 0.27).

Overall, both methods achieve comparable target scores through different acoustic realizations. Best-of-8 selects from the base model's existing distribution of voices and speaking rates, whereas GRPO increases speech rate and, under two of the three rewards, produces a slightly narrower pitch range than Best-of-8.